**REVIEW**

# Graph neural networks in vision-language image understanding: a survey

Henry Senior[1] · Gregory Slabaugh[1] · Shanxin Yuan[1] · Luca Rossi[2]

**Abstract**
2D image understanding is a complex problem within computer vision, but it holds the key to providing human-level scene comprehension. It goes further than identifying the objects in an image, and instead, it attempts to *understand* the scene. Solutions to this problem form the underpinning of a range of tasks, including image captioning, visual question answering (VQA), and image retrieval. Graphs provide a natural way to represent the relational arrangement between objects in an image, and thus, in recent years graph neural networks (GNNs) have become a standard component of many 2D image understanding pipelines, becoming a core architectural component, especially in the VQA group of tasks. In this survey, we review this rapidly evolving field and we provide a taxonomy of graph types used in 2D image understanding approaches, a comprehensive list of the GNN models used in this domain, and a roadmap of future potential developments. To the best of our knowledge, this is the first comprehensive survey that covers image captioning, visual question answering, and image retrieval techniques that focus on using GNNs as the main part of their architecture.

**Keywords** Graph neural networks · Image captioning · Visual question answering · Image retrieval

## 1 Introduction

Recent years have seen an explosion of research into graph neural networks (GNNs), with a flurry of new architectures being presented in top-tier machine learning conferences and journals every year [1–7]. The ability of GNNs to learn in non-Euclidean domains makes them powerful tools to analyse data where structure plays an important role, from chemoinformatics [8] to network analysis [9]. Indeed, these models can also be applied to problems not traditionally associated with graphs such as 3D object detection in LiDAR point clouds [10] and shape analysis [11].

GNN-based approaches have gained increasing popularity for solving 2D image understanding Vision-Language tasks, similar to other domains [12–14]. Whilst advances in this domain are discussed in [15], it is a wide ranging survey. Our work focuses specifically on Vision-Language and therefore covers these topics more extensively.

We view 2D image understanding as the high-level challenge of making a computer understand a two-dimensional image to a level equal to or greater than a human. Models that enable this should be able to reason about the image in order to describe it (image captioning), explain aspects of it (visual question answering (VQA), or find similar images (image retrieval). These are all tasks that humans can do with relative ease; however, they are incredibly difficult for deep learning models and require a large amount of data. These tasks also fall under the category of Vision-Language problems, as they require the model to have an understanding of both the image pixels and a language (typically English) in which the models can express their understanding. Adjacent to these is the challenging task of Vision-Language Navigation [16–21], i.e. the development of a system able to navigate a previously unseen environment using natural language instructions from

✉ Henry Senior
h.senior@qmul.ac.uk

Gregory Slabaugh
g.slabaugh@qmul.ac.uk

Shanxin Yuan
shanxin.yuan@qmul.ac.uk

Luca Rossi
luca.rossi@polyu.edu.hk

[1] Digital Environment Research Institute, Queen Mary University London, New Road, London E1 1HH, UK

[2] Department of Electrical and Electronic Engineering, The Hong Kong Polytechnic University, Hung Hom, Kowloon, Hong Kong

a human and its own visual observations. Combining Vision, NLP, Agents, and potentially Robotics, it is a task that merges together a broad set of fields. Given the task breadth and depth, and its strong links to Agents and Robotics, it falls outside the scope of this survey. Readers are directed to the recent survey by Wu et al. [22] for in-depth reviews of this task. GNNs also have a wide range of applications within the medical imaging domain [23–25], including the generation of medical reports through the utilisation of knowledge graphs [23–25] (an extension of the image captioning task). They have also been used in medical VQA [26] and medical image retrieval [27].

Whilst there is a plethora of techniques that have been applied to the tasks discussed in this survey [28–37], this survey focuses on graph-based approaches. There are a range of graphs that are applicable, but the most widely used and understood is the semantic graph [38, 39]. This graph is constructed of nodes representing visual objects and edges representing the semantic relationships between them. The semantic graph as well as further graph types are discussed in Sect. 2.3.

Alongside a taxonomy of the graph types used across 2D image understanding tasks, this paper contributes a much needed overview of these approaches. Covering the three main tasks, we also include an overview of popular GNN techniques as well as insights on the direction of future GNN work. In the discussion section of this paper, we argue that the increasingly popular Transformer architecture [40] is actually a special case GNN [41]. We expand upon this argument to suggest that GNNs should not be overlooked as they may offer better inductive biases for a range of tasks.

Our main contributions are: (1) a taxonomy of the graph types used in 2D image understanding tasks; (2) a comprehensive survey of GNN-based approaches to common 2D image understanding tasks; and (3) a roadmap of potential future developments for the community to explore.

The remainder of this paper is organised as follows: Sect. 2 gives a taxonomy of the tasks discussed and their corresponding datasets, as well as an overview of the different graph types used throughout. Section 3 gives an overview of the common GNN architectures used. It also briefly mentions current and future research directions for GNNs and signposts appropriate surveys. The main body of the paper is formed of Sects. 4, 5, and 6, which detail GNN-based approaches to image captioning, VQA, and image retrieval, respectively. We then conclude the paper with a three part discussion, with Sect. 7.1 covering the advantages that GNNs still offer despite the rapid adoption of the Transformer architecture. This is followed by Sect. 7.2 which links the emerging field of latent diffusion and image generation to image captioning. Finally, Sect. 7.3 concludes the paper and provides potential directions for future work.

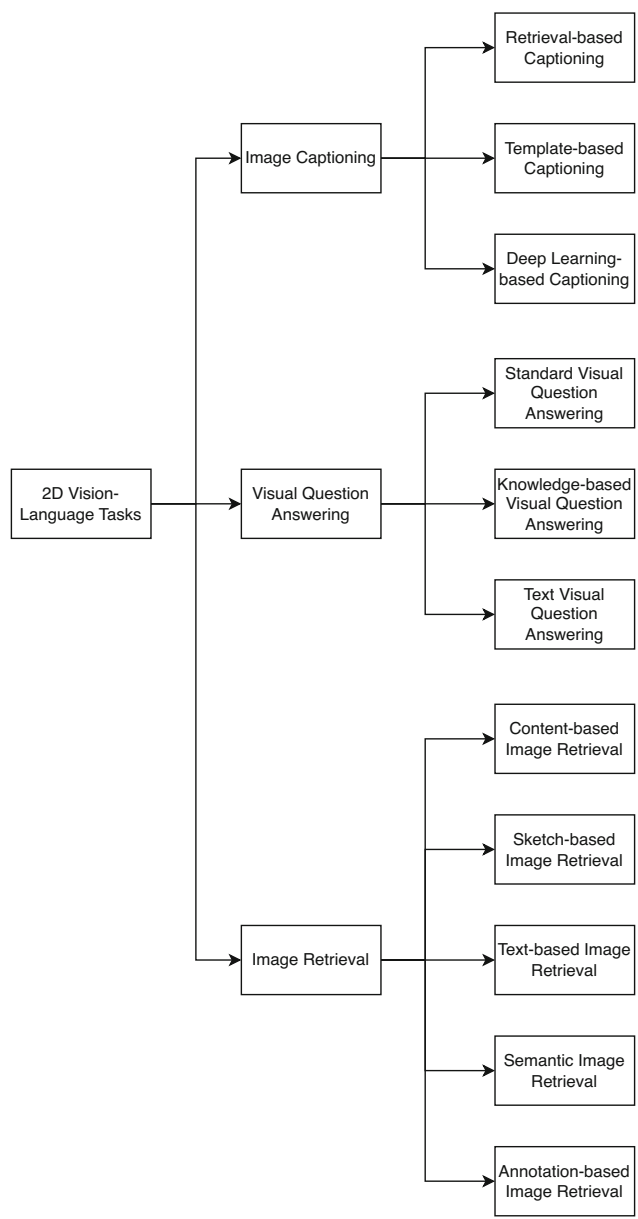

**Fig. 1** 2D vision-language methodological taxonomy

## 2 Background and definitions

In this section, we outline the background required to view this survey in context. We first briefly define a generic graph before outlining the taxonomy of the field. Finally, we give an overview of the various graph types.

### 2.1 2D vision-language tasks taxonomies

This paper follows the taxonomies of [42–45] and joins them together for a more complete overview of 2D Vision-Language tasks (see Fig. 1). This section gives a brief overview of the existing taxonomies and highlights the sec-tions of them this survey focuses on. It also highlights the main datasets used for various tasks discussed in the paper, and these are summarised in Table 1.

Whilst individual Vision-Language tasks have their own unique datasets, they are unified by the Visual Genome [46], an expansive dataset that provides ground truths for a range of Vision-Language tasks. As the most generic dataset, it has 33, 877 object categories and 68, 111 attribute categories. At the time of its publication, this was the largest and most dense dataset containing image descriptions, objects, attributes, relationships, and question–answer pairs. Additionally, the Visual Genome also contains region graphs, semantic graphs, and question–answer pairs. This results in it being a very wide ranging dataset with lots of applications in visual cognition tasks such as scene graph generation [54] and VQA [55].

For image captioning, we follow [42] who identify three main approaches: 1) retrieval-based captioning, the task of mapping an input image to an existing caption; 2) template-based captioning, using image features to complete a caption template; and 3) deep learning-based captioning, where the caption is generated from scratch. Retrieval-based captioning is built on the assumption that for every image, a caption exists, and needs to be retrieved from a bank of existing captions. It was the foundation of early image captioning approaches [28] and yielded good results without the need for deep learning. However, not all images may have appropriate captions. If the captions are generic, they will only be able to describe aspects of an image and may omit its most important feature. In contrast, template-based captioning [56] uses a predefined caption format and uses object detection to fill in the blanks. This approach is good for generating consistent captions, but can result in captions that are unnatural and clearly generated by a machine. Contemporary approaches to the task of image captioning are based on deep learning models. Early work focused on a CNN encoder feeding an RNN-based decoder [57]; however, more recent deep learning approaches have developed to incorporate a wide variety of techniques including GNNs [39, 58] and Transformers [59, 60]. In this survey, we focus specifically on deep learning approaches to image captioning and focus on graph-based approaches. Deep learning approaches are typically trained on the COCO [47] or Flickr30k [48] which contain a set of images accompanied by five human generated captions. Closely related to contemporary deep learning-based captioning are the tasks of paragraph captioning and video captioning. Paragraph image captioning is the challenge of generating a multi-sentence description of an image [61, 62], whilst video captioning focuses on describing videos. Readers interested in video captioning are directed to the recent survey [63].

Taxonomies of VQA are usually defined through the lens of the datasets used by the various tasks [43, 44]. Here we focus on 1) the standard VQA task of answering a ques- tion about an image, 2) the fact-based VQA (FVQA) task of answering questions that require external knowledge to answer, and 3) TextVQA, the task of answering questions that require the model to read text in the scene and combine it with visual data. Each of the various VQA tasks have their own set of speicalised datasets. The original VQA dataset [49] and the subsequently updated VQA 2.0 [64] dataset address the original task of answering questions based on the visual information in the image. The FVQA dataset [50] is built using images from ImageNet [65] and COCO [47] alongside facts from DBPedia [66], ConceptNet [67], and WebChild [68]. The images have three forms of visual concepts extracted from them using a range of models. These visual concepts include objects (items identified in the image), scene (scene-level features such as room label), and actions. Question–answer pairs were generated by human annotators who selected a visual concept and an accompanying fact triplet which they used to generate a question. Finally, the text-KVQA dataset [53] was built by compiling images from a Kaggle movie poster challenge,[1] [69], and Google Image search results from combining brand names with postfixes such as 'store' or 'building'. This collection of images was then given to human annotators who removed images that did not contain text of brand names. The result is a dataset of 257K images with three groupings: book, movie, and scene. Accompanying these images are 1.3 million question–answer pairs. Each image grouping gets its own triplet-based knowledge base from a relevant source: WikiData [70], IMBd, and [69], respectively. Adjacent to the task of VQA and its sub-tasks is the field of Visual Grounding (sometimes known as Referring Expression). This is the task of identifying the salient regions of an image based on a natural language query. Although the task is closely aligned with VQA, it falls outside the scope of this paper. We direct the readers to [15].

Image retrieval spans multiple tasks, all of which make use of deep learning in contemporary approaches. We follow the taxonomy of Alexander et al. [45] and address the following sub-tasks: 1) text-based image retrieval where images are returned based on a text query; 2) content-based image retrieval where images are retrieved based on their similarity to an input image; 3) sketch-based retrieval, where images are retrieved based on their similarity to a sketch; 4) semantic-based retrieval which returns images based on their perceptual content; and 5) annotation-based retrieval, where images are returned using meta-data annotations. The number of datasets used for image retrieval are vast, and the community has not solidified around a single dataset in the way image captioning has around COCO [47]. This presents a challenge when making accurate comparisons between sys-tems as the challenge presented by different datasets varies complicating direct comparisons across datasets. Whilst image retrieval specific datasets exist [71], there are papers [72–74] that make use of image captioning datasets [47, 48], showing the wide range of varied datasets that exist for image retrieval.

---

[1] https://www.kaggle.com/datasets/neha1703/movie-genre-from-its-poster

**Table 1** A summary of common datasets

| Dataset | Main task | Features |
| --- | --- | --- |
| Visual Genome [46] | Multi-Task | 108,000 images; 5.4 million region descriptions; 1.7 million question–answer pairs; scene graphs |
| COCO [47] | Image Captioning | 330,000 images with 5 human generated reference captions for training and validation sets |
| Flickr30K [48] | Image Captioning | 31,000 images each with 5 human generated reference captions |
| VQA 2.0 [49] | VQA | 265,000 images; average of 5.4 questions per image each with 10 ground truth answers |
| FVQA [50] | VQA | 2,190 images; 5,826 questions; knowledge base of 4,216 facts |
| OK-VQA [51] | VQA | 14,000 questions, each with five ground truth answers, with knowledge extracted from Wikipedia |
| TextVQA [52] | VQA | 28,000 images; 45,000 questions each with 10 ground truth answers |
| Text-KVQA [53] | VQA | 257,000 images; 1.3 million QA pairs; inclusion of a knowledge base |

Understanding the inherent biases in datasets is incredibly important for deep learning researchers and practitioners. As models move beyond research benchmarks and into mainstream use, models that are trained on biased data will produce biased outputs and may contribute to the proliferation of harmful stereotypes. Within the scope of Vision-Language, work has been done to discover negative biases in core datasets such as COCO, enabling researchers to mitigate these risks [75]. Comprehensively demonstrated the gender, racial, and Western biases that exist in the COCO [47] dataset. The research finds lighter-skinned individuals are $7.5\times$ more common than darker-skinned individuals, and males are $2\times$ more common than females. This leads to the concern that image captioning models may come to see the world as being predominantly occupied by light-skinned males. Worryingly, [75] also find the existence of racial slurs in the ground truth captions which leads to concerns about captioning models producing captions containing this derogatory language.

Hirota et al. [76] continue the work of [75] but focus on VQA datasets. They find evidence of bias in Visual Genome [46] and OK-VQA [51], two widely used VQA datasets. The biases include a reflection of traditional gender stereotypes and a US-centric viewpoint on race and nationality.

In addition to the gender and racial biases in existing Vision-Language datasets, a lot of the datasets are limited by style. The vast majority of datasets contain real world photographs (typically mined from Flickr), which limits models to only understanding photographs. This limitation is most significant in image captioning and image retrieval as the style is a significant component of either the caption or retrieval query. Unless the question asked is specifically about the style of the image, then the impact is somewhat limited for VQA.

### 2.2 Fundamental graph theoretical concepts

*Undirected graph.* We define an undirected graph $G$ to be a tuple of sets $V$ and $E$, i.e. $G = (V, E)$. The set $V$ contains $n$ vertices (sometimes referred to as nodes) that are connected by the edges in the set $E$, i.e. if $v \in V$ and $u \in V$ are connected by an edge then $e_{v,u} \in E$. For an undirected graph, we have that $e_{v,u} = e_{u,v}$.

*Directed graph* A directed graph is a graph where the existence of $e_{v,u}$ does not imply the existence of $e_{u,v}$ as well. Let $A$ be the $n \times n$ binary adjacency matrix such that $\mathbf{A}_{v,u} = 1$ if $e_{v,u} \in E$. Then it follows that $\mathbf{A}$ is asymmetric (symmetric) for directed (undirected) graphs. More in general, $\mathbf{A}$ can be a real-valued matrix, where the value of $\mathbf{A}_{v,u}$ can be interpreted as the strength of the connection between $v$ and $u$.

*Neighbourhood.* The neighbourhood $\mathcal{N}(v)$ of a vertex $v \in V$ is the subset of nodes in $V$ that are connected to $v$. The neighbour $u$ can be either directly connected to $v$, i.e. $(v, u) \in E$, or indirectly connected by traversing $r$ edges from $v$ to $u$. Note that some definitions include $v$ itself as part of the neighbourhood.

*Complete graph.* A complete graph is one (directed or undirected) where for each vertex, there is an edge connecting it to every other vertex in the set $V$. A complete graph is therefore a graph with the maximum number of edges for a given number of nodes.

*Multi-partite graph.* A multi-partite graph (also known as $K$-partite graph) is a graph where the nodes can be separated into $K$ different sets. For scene understanding tasks, this allows for a graph representation where one set of nodes represent objects and another represents relationship between objects.

*Multi-modal graph.* A multi-modal graph is one with nodes that have features from different modalities. This approach is commonly used in VQA where the image and text modalities are mixed. Multi-modal graphs enable visual features to coexist in a graph with word embeddings.

## 2.3 Common graph types in 2D vision-language tasks

This section organises the various graph types used across all three tasks discussed in the survey. Some graphs, such as the semantic and spatial graphs, are used across all tasks [39, 55, 73], whilst others are more domain specific, like the knowledge graph [53, 77]. Figure 2 shows a sample image from the COCO dataset [47] together with various types of graphs that can be used to describe it. This section, alongside the figure, is organised so that graph that represent a single image and graphs that represent portions the dataset are grouped together.

*Semantic graph, multi-partite semantic graph, and textual semantic graph.* Sometimes referred to as a scene graph, a semantic graph (shown in Fig. 2b) is a one that encapsulates the semantic relationships between visual objects within a scene. Across the literature, the terms 'semantic graph' and 'scene graph' are used somewhat interchangeably, depending on the paper. However, in this survey we use the term 'semantic graph' because there are many ways to describe a visual scene as a graph, whereas the 'semantic graph' label is more precise about what the graph represents. Semantic graphs come in different flavours. One approach is to define a directed graph with nodes representing visual objects extracted by an object detector such as Faster-RCNN [78] and edges representing semantic relationships between them. This is the approach of Yao et al. [39], where, using a dataset such as Visual Genome [46], a model predicts the semantic relationships to form edges in the graph. Alternatively, the semantic graph can be seen as a multi-partite graph [58, 79–81] (shown in Fig. 2c), where attribute nodes describe the object nodes they are linked to. They also change the way relationships are represented by using nodes rather than edge features. This yields a semantic graph with three node types: visual object, object attribute, and inter-object relationship. This definition follows that of the 'scene graph' defined by Johnson et al. [38]. Finally, another form of semantic graph exists, the textual semantic graph[58, 82] (shown in Fig. 2d). Unlike visual semantic graphs, textual ones are not generated from the image itself but rather its caption. Specifically, the caption is parsed through the Stanford Dependency Parser [83], a widely used [84, 85] probabilistic sentence parser. Given a caption, the parser will return its grammatical structure, identifying components such nouns, verbs, and adjectives and marking the relationship between them. This is then modified from a tree into a graph, following the techniques outlined in [86].

*Spatial graph.* Yao et al. [39] define a spatial graph (Fig. 2e) as one representing the spatial relationship between objects. Visual objects detected by an object detector form nodes, and the edges between the nodes represent one of 11 predefined spatial relationships that may occur between the two objects. These include inside (labelled '1'), cover (labelled '2'), overlap (labelled '3'), and eight positional relationships (labelled '4'–'11') based on the angle between the centroid of the two objects. These graphs are directional but will not always be complete as there are cases where two objects have a weak spatial relationship and are therefore not connected by an edge in the spatial graph. Guo et al. [80] define a graph of a similar nature known as a geometry graph. It is defined as an undirected graph that encodes relative spatial positions between objects with an overlap and relative distance that meet certain thresholds.

*Hierarchical spatial (Tree).* These graphs build on from the spatial graph but the relationships between nodes focus on the hierarchical nature of the spatial relationship between the detected objects within an image. Yao et al. [87] propose to use a tree (i.e. a graph where each pair of nodes is connected by a single path) to define a hierarchical image representation. An image ($\mathcal{I}$) is first divided into regions using Faster-RCNN [78] ($\mathcal{R} = \{r_i\}_{i=1}^K$) with each region being further divided into instance segmentations ($\mathcal{M} = \{m_i\}_{i=1}^K$). This gives a three-layer tree structure ($\mathcal{T} = (\mathcal{I}, \mathcal{R}, \mathcal{M}, \mathcal{E}_{tree})$, where $\mathcal{E}_{tree}$ is the set of connecting edges) to represent the image, as shown in Fig. 2f. He et al. [60] use a hierarchical spatial graph, with relationships representing 'parent', 'child', and 'neighbour' relationships depending on the intersection over union of the bounding boxes.

*Similarity graph.* The similarity graph (Fig. 2) proposed by Kan et al.[88] (referred to as a semantic graph by the authors) is generated by computing the dot product between two visual features extracted by Faster-RCNN [78]. The dot products are then used to form the values of an adjacency matrix $A$ as the operation captures the similarity between two vectors, the higher the dot product, the closer the two vectors are. Faster-RCNN extracts a set of $n$ visual features, where each feature $x(v)$ is associated to a node $v$ and the value of the edge between two nodes $v$ and $u$ is given by $\mathbf{A}_{u,v} = \sigma\left(x(v)^T \mathbf{M} x(u)\right)$, where $\sigma(\cdot)$ is a nonlinear function and $\mathbf{M}$ is a learnt weight matrix. The authors of [88] suggest that generating the graph this way allows for relationships between objects to be discovered in a data-driven manner, rather than relying on a model trained on a dataset such as the Visual Genome [46].

*Image graphs/K-nearest neighbour graph.* In their 2021 image captioning work, Dong et al. [89] construct an image graph by converting images into a latent feature space by averaging the object vectors output by feeding the image into Faster-RCNN [78]. The $K$ closest images from the training data or search space in terms of $l_2$ distance are then turned into an undirected complete graph, shown in Fig. 2h. This is a similar approach used by Liu et al. [90] with their $K$-nearest neighbour graph.

*Topic graph.* Proposed by Kan et al. [88], the topic graph is an undirected graph of nodes representing topics extracted by GPU-DMM [91]. Topics are latent features representing shared knowledge across the entire caption set. Modelling them as a graph, as shown in Fig. 2i, with edges computed by taking the dot product of the two nodes, allows the modelling of knowledge represented in the captions.

*Region adjacency graph.* A region adjacency graph (RAG) is a graph made up of nodes representing homogeneous segments of the image, with edges representing the connection of adjacent regions. There are different approaches to defining the regions, but patches or superpixels are commonly used. Patches, small equal divisions of the image, are used by Sui et al. [92] whilst superpixels, an unsupervised segmentation approach for clustering nearby pixels, are used by [93].

*Knowledge graph.* A knowledge graph, or fact graph, is a graph-based representation of information. Whilst there is no agreed structure of these graphs [94], they typically take the form of triplets. They are used in a wide variety of tasks to provide the information needed to 'reason'. Hence, knowledge graphs enable the FVQA task.

## 3 An overview of graph neural networks

Over the past years, a large number of GNN architectures have been introduced in the literature. Wu et al. [95] proposed a taxonomy containing four distinct groups: recurrent GNNs, convolutional GNNs, autoencoder GNNs, and spatial–temporal GNNs. The applications discussed in this paper mostly utilise convolutional GNNs, for a comprehensive overview of other architectures readers are directed to [95]. GNNs, especially traditional architectures such as graph convolutional network, have a deep grounding in relational inductive biases [41]. They are built on the assumption of homophily, i.e. that connected nodes are similar. There is an increasing body of work looking into addressing some of the bottlenecks that GNNs may suffer from. Novel training strategies such as [96] have been shown to reduce GPU memory whilst approaches such as [97] reduce the difference in performance when dealing with homophilic or heterophilic graphs.

### 3.1 Graph convolutional networks

One common convolutional GNN architecture is the message passing neural networks (MPNNs) proposed by Gilmer et al. Although this architecture has been shown to be limited [98], it forms a good abstraction of GNNs.

Gilmer et al. describe MPNNs as being comprised of a message function, update function, and readout function. These functions will vary depending on the application of the network, but are learnable, differentiable, and permutation invariant. The message and update functions will run for a number of time steps $T$, passing messages between connected nodes of the graph. These are used to update the hidden feature vectors of the nodes, which are then used to update the node feature vector, which in turn is used in the readout function.

The messages are defined as

$$\mathbf{m}_v^{(t+1)} = \sum_{u \in \mathcal{N}(v)} M_t(\mathbf{h}_v^{(t)}, \mathbf{h}_u^{(t)}, \mathbf{e}_{v,u}), \tag{1}$$

where a message for a node at the next time step $\mathbf{m}_v^{(t+1)}$ is given by combining its current hidden state $\mathbf{h}_v^{(t)}$ with that of its neighbour $\mathbf{h}_u^{(t)}$ and any edge feature $\mathbf{e}_{v,u}$ in a multi-layer perceptron (MLP) $M_t(\cdot)$. Given that a message is an aggregation of all the connected nodes, the summation acts over the nodes connected to the node $u \in \mathcal{N}(v)$, i.e. the neighbourhood of $v$.

These messages are then used to update the hidden vectors by combining the node current state with the message in an MLP $U_t$.

$$\mathbf{h}_v^{(t+1)} = U_t(\mathbf{h}_v^t, \mathbf{m}_v^{(t+1)}) \tag{2}$$

Once the message passing phase has run for $T$ time steps, a readout phase is then conducted using a readout function, $R(\cdot)$. This is defined as an MLP that considers the updated feature vectors of nodes ($\mathbf{h}_v^T$) on the whole graph ($v \in G$) to produce a prediction and is defined as:

$$\hat{y} = R(\{\mathbf{h}_v^T | v \in G\}) \tag{3}$$

In order to make the GCN architecture scale to large graphs, the GraphSAGE [99] architecture changes the message function. Rather than taking messages from the entire neighbourhood of a node, a random sample is used. This reduces the number of messages that require processing, resulting in an architecture that works well on large graphs.

### 3.2 Gated graph neural networks

The core idea behind the gated graph neural network (GGNN) [100] is to replace the update function from the

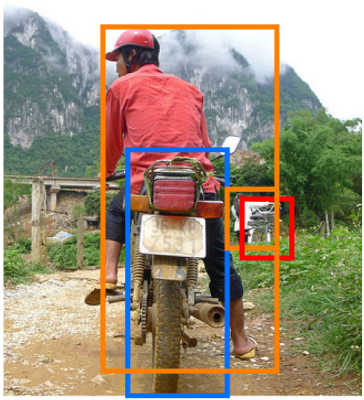

(a) COCO training image 391895 (Cropped) with object detection boundaries

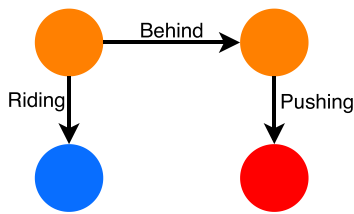

(b) A semantic graph where detected objects are represented by nodes, and the edges are predicted by a model trained on the Visual Genome 'scene graph' generation task [46].

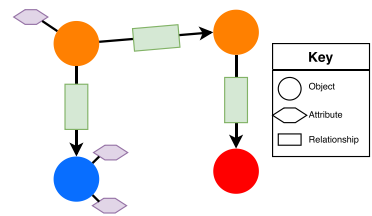

(c) A multipartite semantic graph builds on the traditional scene graph (Figure 2b) by representing the relationships as nodes and adds additional nodes to represent visual attributes.

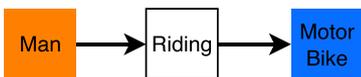

(d) A textual semantic graph representation of the caption 'A man riding on the back of a motor bike'

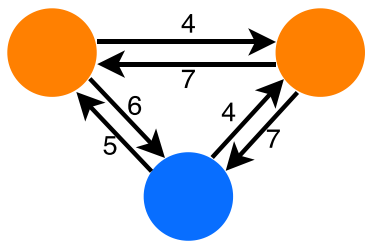

(e) A spatial graph representation of an image, where the nodes represent the objects in the image and the edges represent the different spatial relationships between the object bounding boxes.

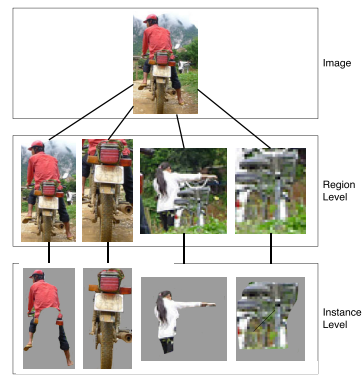

(f) An image tree representation, where objects are divided into region level and instance level attributes.

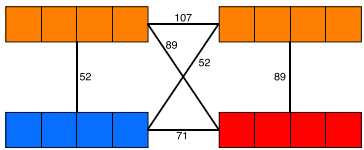

(g) A similarity graph representation. In a similarity graph, the dot product between image features is calculated and stored in an adjacency matrix.

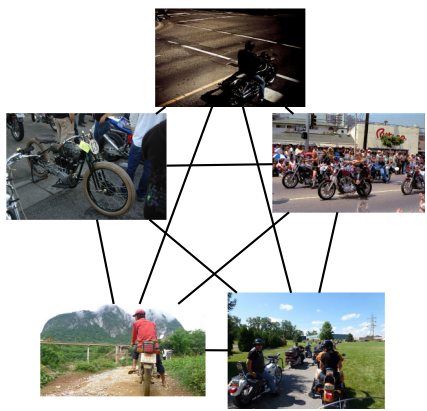

(h) An example image graph representation of the wider COCO dataset. Edges connect similar images together - making this graph representative of the wider dataset rather than an individual image.

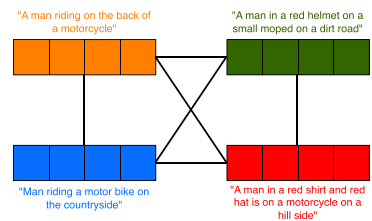

(i) A topic graph representation, where the nodes are embeddings of topics extracted from the caption set. The dot product is calculated between the embeddings and stored in the adjacency matrix.

**Fig. 2** A visual comparison of the various graph types used across vision-language tasks. Best viewed in colour

message passing architecture (Eq. 2) with a gated recurrent unit (GRU) [101]. The GRU is a recurrent neural network with a update and reset gates that controls which data can flow through the network (and be retained) and which data cannot (and therefore be forgotten).

$$\mathbf{h}_v^{(t+1)} = GRU\left(\mathbf{h}_v^{(t)}, \sum_{u \in \mathcal{N}(v)} \mathbf{W}\mathbf{h}_u^{(t)}\right). \quad (4)$$

where $\mathbf{h}_v^t$ is the hidden feature of node $v$ at time $t$, $\mathbf{W}$ is a learnt weight matrix, and $u \in \mathcal{N}(v)$ is the subset of nodes in the graph connected to node $v$.

The GGNN also replaces the message function from Eq. 1 with a learnable weight matrix. Using the GRU alongside back-propagation through time enables the GGNN to operate on series data. However, due to the recurrent nature of the architecture, it can become unfeasible in terms of memory to run the GGNN on large graphs.

### 3.3 Graph attention networks

Following on from the multi-head attention mechanism of the popular Transformer architecture [40], graph attention networks (GATs) [102] extend the common GCN to include this attention attribute. Using an attention function, typically modelled by an MLP, the architecture calculates an attention weighting between two nodes. This process is repeated $K$ times using $K$ attention heads in parallel. The attention scores are then averaged to give the final weights.

The self-attention is computed by a function $a(h_v^t, h_u^t)$ (typically an MLP) that attends to the hidden representation of a node ($h_v^t$) and one of its neighbours ($h_u^t$). Once every node pairing in the graph has their attention computed, the scores are passed through a softmax function to give a normalised attention coefficient ($\alpha_{v,u}$). This process is then extended to multi-head attention by repeating the process across $K$ different attention heads, each with different initialisation weights. The final node representation is achieved by concatenating or averaging (represented as $\|$) the $K$ attention heads together.

$$\mathbf{h}_v^{(t+1)} = \Big\|_{k=1}^{K} \sigma\left(\sum_{u \in \mathcal{N}(v)} \alpha_{v,u}^{(k)} \mathbf{W}^{(k)} \mathbf{h}_u\right) \quad (5)$$

where $\sigma$ is a nonlinear activation function such as ReLU and $\mathbf{W}$ is a learnable weight matrix.

### 3.4 Graph memory networks

Recent years have seen the development of graph memory networks, which can conceptually be thought of as models with an internal and external memory. When there are multiple graphs overlapping the same spatial information, as in [103], the use of some form of external memory can allow for an aggregation of node updates and the graph undergoes message passing. This essentially allows for features from multiple graphs to be combined in some way that goes beyond a more simplistic pooling operation. In the case of Khademi [103], two graphs are constructed across the same image but may have different nodes. These graphs are updated using a GGNN. An external spatial memory is constructed to aggregate information from across the graphs as they are updated, using a neural network with an attention mechanism. The final state of the spatial memory is used to perform the final task.

### 3.5 Modern graph neural network architectures

In recent years, the limits of message passing GNNs have become increasingly evident, from their tendency to over-smooth the input features as the depth of the network increases [104] to their unsatisfactory performance in heterophilic settings [105], i.e. when neighbouring nodes in the input graphs are dissimilar. Furthermore, the expressive power of GNNs based on the message passing mechanism has been shown to be bounded by that of the well-known Weisfeiler–Lehman isomorphism test [98], meaning that there are inherent limits to their ability to generate different representations for structurally different input graphs.

Motivated by the desire to overcome these issues, researchers have now started looking at alternative models that move away from standard message passing architectures. Efforts in this direction include, among many others, higher-order message passing architectures [106], cell complexes networks [107], networks based on diffusion processes [2, 105, 108]. To the best of our knowledge, the application of these architectures to the 2D image understanding tasks discussed in this paper has not been explored yet. As such, we refer the readers to the referenced papers for detailed information on the respective architectures.

## 4 Image captioning

Image captioning is the challenging task of producing a natural language description of an image. Outside of being an interesting technical challenge, it presents an opportunity to develop accessibility technologies for severely sight impaired (formally 'blind') and sight impaired users (formally 'visually impaired' [2]). Additionally, it has applications in problems ranging from image indexing [109] to surveillance [88]. There are three forms of image captioning techniques: 1) retrieval-based captioning, where a caption is retrieved from a set of existing captions, 2) template-based captioning, where a pre-existing template is filled in using information extracted from the image, and 3) deep learning-based image captioning, where a neural network is tasked with generating a caption from an input image. We propose to refine this taxonomy to differentiate between GNN-based approaches and more traditional deep learning powered image captioning. The following section details the GNN-based approaches to image captioning, of which there have been a number of in recent years. Figure 3 illustrates the structure of a generic GNN-based image captioning architecture.

---

[2] The UK Department of Health and Social Care adopted the more inclusive phrasing around 2017.

GNN-based approaches to image captioning all follow the traditional encoder–decoder-based approach common in deep learning image captioning techniques. Images first undergo object detection, the output of which is used to create an encoding. These encodings are then decoded, traditionally with a long short-term memory network (LSTM), into a caption. Through incorporating GNNs, researchers have been able to enhance the encoded image representation by incorporating spatial and semantic information into the embeddings.

As the task of image captioning has developed over time, so have the evaluation metrics used to assess the performance of proposed architectures. Originally, image captioning relied heavily on machine translation evaluation techniques such as BLEU [110], ROUGE [111], and METEOR [112] as no image captioning specific metric existed. However, this changed with the introduction of both CIDEr [113] and SPICE [86]. The performance metrics are detailed in Table 2.

The first architecture to use a GNN to improve image captioning was by Yao et al. [39]. In their work, they propose the use of a GCN to improve the feature embeddings of objects in an image. They first start by applying a Faster RCNN object detector [78] to the image in order to extract feature vectors representing objects. These feature vectors are then used to create two graphs: a bidirectional spatial graph encoding spatial relationships between objects and a directed semantic graph which encodes the semantic relationships between objects. A GCN is then applied to both graphs before the enhanced features of the graphs undergo mean pooling. They are then decoded by an LSTM into a caption. As the whole graphs are used to inform the caption generation, it may lead to scenarios where dense graphs lead to redundant or low-value information being included in the caption.

Zhong et al. [79] focus solely on a semantic graph and address the problem of which nodes and edges to include in the final caption. This is challenging for scenes containing a lot of detected objects as the semantic graphs can become relatively large. The problem is addressed by decomposing the semantic graph into various sub-graphs that cover various parts of the image. They are then scored using a function trained to determine how closely the sub-graph resembles the ground truth caption. This enables the selection of sub-graphs from the main semantic graph (produced by the commonly used MotifNet [114]) that will go on to generate useful captions. Zhong et al. [79] make use of a GCN to aggregate neighbourhood information of the proposed sub-graph; focusing on the link between the language modality and semantic graph, therefore discarding the spatial information.

Another work that makes use of the semantic graph is that of Song et al. [115]. They investigated how both implicit and explicit features can be utilised to generate accurate and high-quality image captions. The authors define implicit features as representing global interactions between objects and explicit features as those defined on a semantic graph. For the latter, rather than using multiple graphs, [115] only uses a single semantic graph. However, rather than predicting the graph directly via MotifNet [114] as in other works [79], its construction starts with a spatial graph. After object detection, a fully connected directed graph is generated between the objects (with nodes being represented by the object feature vector). The edges of this graph are then whittled away in a two step process. Firstly, edges between objects that have zero overlap (measured as intersection over union) and an $l_2$ distance less than the longest side of either objects bounding box are removed. The remaining edges are used to determine which object pairs have their relationship detected by MotifNet [114]. Those relationships with a high enough probability are kept whilst the others are removed. This results in a semantic graph that indirectly contains spatial information, going beyond the semantic graph of [79]. The final graph is then processed by a GGNN, the output of which is a representation of the explicit features. The implicit features are generated by a Transformer encoder [40]. The entire image alongside the regions within the detected object bounding boxes is encoded. These features are then used alongside those of the explicit features as input to an LSTM language decoder that is used to generate the final caption. The work demonstrates the successes possible when using GNNs alongside Transformers, using their different inductive biases to best model different interactions (see Table 3). However, both the implicit and explicit relationships remain local to a single image. Further work could consider how often certain relationships occur over the entire dataset.

Guo et al. [80] took a very similar approach to Yao et al. [39] with their work, utilising a dual graph architecture containing a semantic and spatial graph. However, they make the observation that images can be represented by a collection of visual semantic unit (VSU) vectors, which represent an object, its attributes, and its relationships. These VSUs are combined into a semantic graph that models relationships as nodes rather than edge features and adds attribute nodes con-nected to objects, thus making it multi-partite. Doing so gives the graph a closer resemblance to the captions it will go on to generate as objects map to nouns, relationships to verbs and prepositions, and finally attributes to adjectives. The authors argue that this approach allows the model to explicitly learn relationships and model them directly. As argued in [80], a semantic graph of an image has a close mapping to the image caption. Nodes representing objects map directly to nouns, edge features (in the case of [39]) or nodes (in the case of [80]) that encode relationships map clearly to prepositions, and nodes representing attributes map to adjectives. This strong relationship between the graph structure generated by the encoder and the final sentence outputted by the decoder further supports the use of the image graph–sentence architecture used by many image captioning systems.

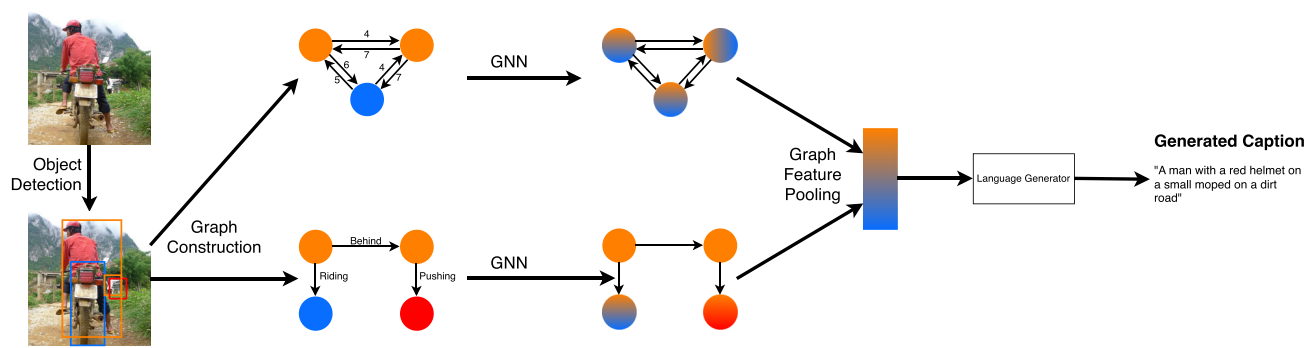

**Fig. 3** An abstract overview of GNN-based image captioning architectures discussed in this section. Most architectures extract image features and use them to construct at least one graph to represent the image. Some papers [88, 89] build higher-level graphs at an image level rather than an object level. A GNN is then applied to these graphs, and the resulting features are fed into a language generator that creates an appropriate caption for the image. Traditionally this was an LSTM, but more recently the trend is to use Transformers [40]. Best viewed in colour

**Table 2** A table detailing the different image captioning performance metrics

| Metric | Original field | Based on | Description |
| --- | --- | --- | --- |
| BLEU | Machine translation | Precision | Based on a modified n-gram precision where the reference word is exhausted after a matching candidate word is identified. BLEU favours captions that are a similar in length to the reference caption |
| ROUGE | Machine text summarisation | Recall | Built with four components: an n-gram recall between the candidate and reference set, a comparison of the longest common sub-sequence, a comparison of the weighted longest common sub-sequence, and finally, the skip-bigram co-occurrence statistic |
| METEOR | Machine translation | $F_{mean}$ | Uses the harmonic mean of the precision and recall between candidate caption and reference captions |
| CIDEr | Image Captioning | n-grams | The metric is based on a number of intuitions. Firstly, that if an n-gram is not present in the reference captions it should not appear in the candidate caption. Secondly, that it should encode how often n-grams present in the candidate captions are present. And finally, n-grams that occur across all the reference captions should be assigned a lower weighting as they will be things like articles and have little to no important information |
| SPICE | Image Captioning | Textual Semantic Tree | Reference and candidate captions are converted into textual semantic graphs to compare the semantic makeup of the captions |

Zhou et al. [81] use an LSTM alongside a Faster-RCNN [78]-based image feature extractor, with the addition of a visual self-attention mechanism. The authors make use of a multi-partite semantic graph, following the style of [38, 80]. Specifically, they propose to use three GCNs to create context-aware feature vectors for each of the object, attribute, and relationship nodes. The resulting context-aware nodes undergo fusion with the self-attention maps, enabling the model to control the granularity of captions. Finally, the authors test two methods of training an LSTM-based language generator: the first being a traditional supervised approach with cross-entropy loss and the second being a reinforcement learning-based approach that uses CIDEr [113] as the reward function. By utilising context dependent GCNs in their architecture, to specifically account for the object, attribute, and relationship nodes, SASG is able to achieve competitive results when compared with similar models, as shown in Table 3.

SGAE (scene graph autoencoder) is another paper to make use of a multi-partite semantic graph. In the paper, Yang et al. [58] take a caption and convert it into a multi-partite textual semantic graph using a similar process to that of the SPICE metric [86] (detailed further in Table 2). The nodes of the graph are converted to word embeddings which are then converted into feature embeddings by way of a GCN, with each node type being given its own GCN with independent parameters. These feature embeddings are then combined with a dictionary to enable them to be re-encoded before they are used to generate a sentence. The dictionary weights are updated via back-propagating the cross-entropy loss from the sentence regeneration. By including a dictionary, the authors are able to learn inductive biases from the captions. This allows generated captions to go from 'man on motorcycle' to 'man riding motorcycle'. When given an image, SGAE generates a multi-partite visual semantic graph, similar to [38, 80], using Faster-RCNN [78] and MotifNet [114]. These visual features are then combined with their word embeddings through a multi-modal GCN and then re-encoded using the previously learnt dictionary. These features are then used to generate the final sentence.

Yang et al. [116] take a multi-partite semantic graph and input it to a multi-head attention-based GNN. The MHA-GNN is based on the Transformer architecture in that a multi-head self-attention is computed between all the nodes of the graph. However, the output of the self-attention is masked by an adjacency matrix prior to the softmax. Doing so enables a self-attention mechanism that maintains the original semantic graph structure. Additionally, the model makes use of Mixture of Experts (MoE) decoding, a first for image captioning. Each node type (object, relationship, attribute) gets its own decoder, and the output is put through a soft router which computes the final output token.

Rather than utilising multiple graphs, Wang et al. [117] instead use a single fully connected spatial graph with an attention mechanism to learn the relationships between different regions. This graph is formed of nodes that represent the spatial information of regions within the image. Once formed, it is passed through a GGNN [100] to learn the weights associated with the edges. Once learnt, these edge weights correspond to the probability of a relationship existing between the two nodes.

The work of Yao et al. [87], following on from their GCN-LSTM [39], presents an image encoder that makes use of a novel hierarchy parsing (HIP) architecture. Rather than encoding the image in a traditional scene graph structure like most contemporary image captioning papers [39, 79, 89], Yao et al. [87] take the novel approach of using a tree structure (discussed in Sect. 2.3), exploiting the hierarchical nature of objects in images. Unlike their previous work which focused on the semantic and spatial relationships, this work is about the hierarchical structure within an image. This hierarchical relationship can be viewed as a combination of both semantic and spatial information—therefore merging the two graphs used previously. The feature vectors representing the vertices on the tree are then improved through the use of Tree-LSTM [118]. As trees are a special case graph, the authors also demonstrate that their previous work GCN-LSTM [39] can be used to create enriched embeddings from the tree before decoding it with an LSTM. They demonstrate that the inclusion of the hierarchy passing improves scores on all benchmarks when compared with GCN-LSTM [39], which does not use hierarchical relationships.

The work of He et al. [60] build on the idea of a hierarchical spatial relationships proposed by Yao et al.[87]. However, rather than use a tree to represent these relationships, they use a graph with three relationship types: parent, neighbour, and child. They then propose a modification to the popular Transformer layer to better adapt it to the task of image processing. After detecting objects using Faster-RCNN [78], a hierarchical spatial relationship graph is constructed. Three adjacency matrices are then built from this graph to model the three relationship types ($\Omega_p$, $\Omega_n$, $and \Omega_c$, respectively). The authors modify the Transformer layer so that rather compute self-attention across the whole spatial graph, there is a sub-layer for each relationship type. Each sub-layer processes the query $Q$ with its own key $K_i$ and value $V_i$ with the modified attention mechanism:

$$Attention(Q, K_i, V_i) = \Omega_i \odot Softmax\left(\frac{QK_i^T}{\sqrt{d}}\right) V_i \quad (6)$$

where $\odot$ is the Hadamard product and $i$ refers to the relationship type $i \in \{parent, neighbour, child\}$ and is used to specify the adjacency matrix ($\Omega$) used. Finally, $\sqrt{d}$ is used as a regularisation technique with $d$ being the dimension of $K_i$. Using the Hadamard product essentially zeroes out the attention between regions whose relationship is not being processed by that sub-layer. The resulting encodings are decoded by an LSTM to produce captions.

Like [60], the $\mathcal{M}2$ meshed memory Transformer proposed by Cornia et al. [59] also makes use of the increasingly popular Transformer architecture [40]. Unlike other papers [39, 58, 60, 87] which make use of some predefined structure on extracted image features (spatial graph, semantic graph, etc.), $\mathcal{M}2$ uses stacks of self-attention layers across the set of all the image regions. The standard key and values from the Transformer are edited to include the concatenation of learnable persistent memory vectors. These allow the architecture to encode a priori knowledge such as 'eggs' and 'toast' make up the concept 'breakfast'. When decoding the output of the encoder, a stack of self-attention layers is also used. Each decoder layer is connected via a gated cross-attention mecha-nism to each of the encoder layers, giving way to the 'meshed' concept of the paper. The output of the decoder block is used to generate the final output caption.

The work of Herdade et al. [119] modifies the attention weight matrix in order to incorporate relative geometric relationships between detected objects. These geometric relationships are defined using a displacement vector that characterises the difference in geometry between two bounding boxes. The work allows the Transformer-based architecture to incorporate geometric relationships directly into the attention mechanism, a relationship not considered by other Transformer-based image captioning techniques such as [59].

The authors of [88] propose using a novel similarity (referred to as a semantic in the paper) and topic graphs. Built on dot product similarity, the graphs are produced without the requirement of graph extraction models such as MotifNet [114]. Rather, a set of vertices $V = \{v_i \in \mathbb{R}^{d_{obj}}\}_{i=1}^{n_{obj}}$ are extracted as ResNet features from a Faster-RCNN object detector [78]. Edges in the adjacency matrix are then populated using the dot product between the feature vectors in $V$ with $a_{ij} = \sigma(v_i^T \mathbf{M} v_j)$, where $\mathbf{M}$ is a matrix of learnable weights and $\sigma$ is a nonlinear activation function. Once both graphs have been constructed, a GCN is applied to both in order to enrich the nodes with local context. A graph self-attention mechanism is then applied to ensure nodes are not just accounting for their immediate neighbours. The improved graphs are then decoded via an LSTM to generate captions.

Following [39], Dong et al. [89] use a spatial graph to show a directed relationship between detected objects within the input image. Locally, object features are extracted by a CNN to associate a vector to each vertex of the spatial graph. This process is completed for each image in the dataset. In addition to this graph, the authors introduce an image-level graph. Specifically, each image is represented by a feature vector that is the average of its associated set of object feature vectors. The image graph for a corresponding image is formed as a fully connected undirected graph of the $K$ images whose $l_2$ distance is the closest to the input image. Both the local spatial graph and the more global image-level graph are processed by GCNs to create richer embeddings that can be used for caption generation. This approach is shown to work extremely well, with Dual-GCN achieving outperforming comparable models in the BLEU, METEOR, and ROGUE metrics (see Table 3).

## 5 Visual question answering

VQA is the challenging task of designing and implementing models that are able to answer natural language questions about a given image. These answers can range from simple yes/no to more natural, longer form answers. Questions can also vary in complexity. As the field has developed, more specific VQA tasks have emerged. The first to emerge was FVQA, sometimes known as knowledge visual question answering (KVQA), where external knowledge sources are required to answer the questions. Another task that has emerged is Textual VQA, where the models must understand the text within the scene in order to generate answers. All three tasks have their own datasets [46, 49, 50, 52, 53] and have an active community developing solutions [49, 84, 103].

### 5.1 VQA

Originally proposed in [49], VQA has developed beyond simple 'yes' or 'no' answers to richer natural language answers. A common thread of work is to leverage the multi-modal aspect of VQA and utilise both visual features from the input image and textual features from the question [84, 85, 103].

One of the first works in VQA to make use of GNNs was that of Teney et al. [84]. Their work is based on the clip art focused dataset [49]. Their model takes a visual scene graph as input alongside a question. The question is then parsed into a textual scene graph using the Stanford Dependency Parser [83]. These scene graphs are then processed independently using a GGNN [100] modified to incorporate an attention mechanism. The original feature vectors are then combined using an attention mechanism that reflects how relevant two nodes from the scene graphs are to one another.

Khademi [103] takes a multi-modal approach to VQA by using dense region captions alongside extracted visual features. Given a query and input image, the model will first extract visual regions using a Faster-RCNN object detector and generated a set of features using ResNet and encoding the bounding box information into these features. An off-the-shelf dense region captioning model is also used to create a set of captions and associated bounding boxes. The captions and bounding box information are encoded using a GRU. Each set of features is turned into a graph (visual and textual, respectively) with outgoing and incoming edges existing between features if the Euclidean distance between the centre of the normalised bounding boxes is less than $\gamma = 0.5$. Both graphs are processed by a GGNN with updated features being used to update an external spatial memory unit—thus making the network a graph memory network (described in Sect. 3.4). After propagating the node features, the final state of the external spatial memory network is turned into a complete graph using each location as a node. This final graph is processed by a GGNN to produce the final answer. The multi-modal approach presented in this paper is shown to be highly effective when compared to similar VQA methods. This approach is shown to work extremely well in benchmarks, with the proposed MN-GMN architecture [103] performing favourably with comparable models (Table 4).

**Table 3** A table showing the model details and various benchmark results of selected image captioning models trained on the COCO [47] dataset using the Karpathy split [57]

| Model | Graph Types | Architecture(s) | Language Generator | BLEU-1 | BLEU-4 | METEOR | ROGUE | CIDEr | SPICE |
|---|---|---|---|---|---|---|---|---|---|
| ARL [117] | Spatial | GGNN | LSTM | 75.9 | 35.8 | 27.8 | 56.4 | 111.3 | – |
| HIP [87] | Hierarchical Spatial (Tree) | GCN | LSTM | – | 39.1 | 28.9 | 59.2 | 130.6 | 22.3 |
| Image Transformer [60] | Hierarchical Spatial | Transformer | LSTM | 80.8 | 39.5 | 29.1 | 59.0 | 130.8 | 22.8 |
| Dual-GCN [89] | Spatial, Image | (Dual) GCN | Transformer | **82.2** | 39.7 | 29.7 | **59.7** | 129.2 | – |
| SUB-GC [79] | Semantic | GCN | LSTM | 76.8 | 36.2 | 27.7 | 56.6 | 115.3 | 20.7 |
| SASG [81] | Bipartite Semantic | GCN | LSTM | 81.8 | 38.9 | 29.2 | 59.4 | 128.9 | **25.0** |
| GCN-LSTM [39] | Spatial, Semantic | GCN | LSTM | 80.9 | 38.3 | 28.6 | 58.5 | 128.7 | 22.1 |
| EIVRN [115] | Spatial, Semantic | GGNN | Transformer | – | 39.4 | 29.3 | 59.1 | 131.9 | 22.8 |
| VSUA [80] | Multi-partite Semantic, Spatial | GCN | LSTM | – | 38.4 | 28.5 | 58.4 | 128.6 | 22.0 |
| SGAE [58] | Multi-partite Textual, Multi-partite Semantic | GCN, multi-modal GNN | LSTM | 80.8 | 38.4 | 28.4 | 58.6 | 127.8 | 22.1 |
| TFSGC [116] | Multi-partite Semantic | GCN | MoE Transformer | – | **39.9** | **29.8** | 59.4 | **133.0** | 23.4 |
| Topic [88] | Similarity | GCN | LSTM | – | 39.2 | 29.1 | 59.0 | 129.5 | 22.6 |
| $\mathcal{M}^2$ [59] | – | Transformer | Transformer | 80.8 | 39.1 | 29.2 | 58.6 | 131.2 | 22.6 |
| Object Relation Transformer [119] | – | Transformer | Transformer | 80.5 | 38.6 | 28.6 | 58.4 | 128.3 | 22.6 |

Bold: Best score

MORN [85] is another work that focuses on capturing the complex multi-modal relationships between the question and image. Like many recent works in deep learning, it adopts the Transformer [40] architecture. Built with three main components, the model first creates a visual graph of the image starting from a fully connected graph of detected objects and a GCN is used to aggregate the visual features. The second part of the model creates a textual scene graph from the input question. Both graphs are merged together by the final component of the model, a relational multi-modal Transformer, which is used to align the representations.

Sharma et al. [120] follow the Vision-Language multi-modal approach but diverge from the use of a textual semantic graph and instead opt to use word embeddings. The authors utilise a novel GGNN-based architecture that processes an undirected complete graph of nodes representing visual features. Nodes are weighted with the probability that a relationship occurs between them. In line with other VQA work [103], the question is capped to 14 words, with each one being converted into GloVe embeddings [121]. Questions with fewer than 14 words are padded with zero vectors. A question embedding is then generated using a GRU applied to the word embeddings. An LSTM-based attention mechanism considers both the question vector and the visual representations making up the nodes of the scene graph. This module considers previously attended areas when exploring new visual features. Finally, an LSTM-based language generator is used to generate the final answer. Another work to forgo using a textual scene graph, Zhang et al. [55] make use of word vectors to embed information about the image into a semantic graph. Using a GNN, they are able to create enriched feature vectors representing the nodes, edges, and an image feature vector representing the global state. They include the question into the image feature by averaging the word vectors, which enables the GNN to reason about the image. Whilst both [120] and [55] yield good results, by only using word- or sentence-level embeddings and not using a textual scene graph, they fail to model relationships in the textual domain. This therefore removes the ability for the models to reason in that domain alone.

Both Li et al. [122] and Nuthalapati et al. [123] take a different route to the established multi-modal approach and instead use different forms of visual information. Li et al. [122] take inspiration from [39] and make use of both semantic and spatial graphs to represent the image. In addition to these explicit graphs, they also introduce an implicit graph, i.e. a fully connected graph between the detected objects with edge weights set by a GAT. The relation-aware visual features are then combined with the question vector using multi-modal fusion. The fused output is then used to predict an answer via an MLP.

Nuthalapati et al. [123] use a dual scene graph approach, using both visual and semantic graphs. These graphs are merged into a single graph embedding using a novel GAT architecture [102] that is able to attend to edges as well as nodes. The graphs are enriched with negative entities that appear in the question but not the graph. Pruning then takes place to remove nodes and edges that are $K$ hops away from features mentioned in the question. A decoder is then used to produce an answer to the inputted question.

### 5.2 Knowledge-/fact-based VQA

Knowledge- or fact-based VQA is the challenging task of making use of external knowledge given in knowledge graphs such as WikiData [70] to answer questions about an image. The major challenge of this task is to create a model that can make use of all three mediums (image, question, and fact) to generate an appropriate answer. The MUCKO [124] architectural diagram shown in Fig. 4 (reused with permission), is shown as a representative example of models that approach FVQA.

In [125], the authors present a novel GCN-based architecture for FVQA. Alongside the question and answer sets, a knowledge base of facts is also included, $KB = \{f_1, f_2, ..., f_{|KB|}\}$. Each fact $f = (x, r, y)$ is formed of a visual concept grounded in the image $(x)$, an attribute or phrase $(y)$, and a relation linking the two $r$. Relationships exist in a predefined set of 13 different ways a concept and attribute can be related. Their work first reduces the search space to the 100 facts most likely to contain the correct answer by using GloVe embeddings [121] of words in the question and facts before further reducing it to the most relevant facts $f_{rel}$. These most relevant facts are turned into a graph where all the visual concepts and attributes from $f_{rel}$ form the nodes. An edge joins two nodes if they are related by a fact in $f_{rel}$. A GCN is then used to 'reason' over the graph to predict the final answer. Using a message passing architecture, the authors are able to update the feature representations of the nodes which can then be fed into an MLP which predicts a binary label corresponding to whether or not the entity contains the answer.

Zhu et al. [124] use a multi-modal graph approach to representing images with a visual, semantic, and knowledge graph. After graph construction, GCNs are applied to each modality to create richer feature embeddings. These embeddings are then processed in a cross-modal manner. Visual–fact aggregation and semantic–fact aggregation operations produce complimentary information which is then used with a fact–fact convolutional layer. This final layer takes into account all three modalities and produces an answer that considers the global context. The authors continue their work in [77] by changing the cross-modal mechanism for a novel GRUC (Graph-based Read, Update, and Control) mechanism. The GRUC operates in a parallel pipeline. One pipeline starts with a concept from the knowledge graph and recurrently incor-

**Table 4** A table showing the model details and VQA [49] Test-Dev results of selected VQA models

| Model | Graphs Used | Architecture | Overall | Y/N | Number | Other | Test-Std |
|---|---|---|---|---|---|---|---|
| Sharma et al [120] | Semantic | GGNN | 67.96 | 84.12 | 46.12 | 58.13 | 67.98 |
| GraphVQA [84] (Abstract Scenes only) | Visual and Textual Semantic | GCN | 70.42 | 81.26 | **76.47** | 56.28 | – |
| MORN [85] | Visual and Textual Semantic | GCN | 71.21 | 87.15 | 55.22 | 61.19 | 71.53 |
| MN-GMN [103] | Visual and Textual Semantic | Graph Memory Network | **73.2** | **88.2** | 56 | **64.2** | **73.5** |
| ReGAT [122] | Semantic, Spatial | GAT | 70.27 | 86.08 | 54.42 | 60.33 | 70.59 |

Bold: Best score

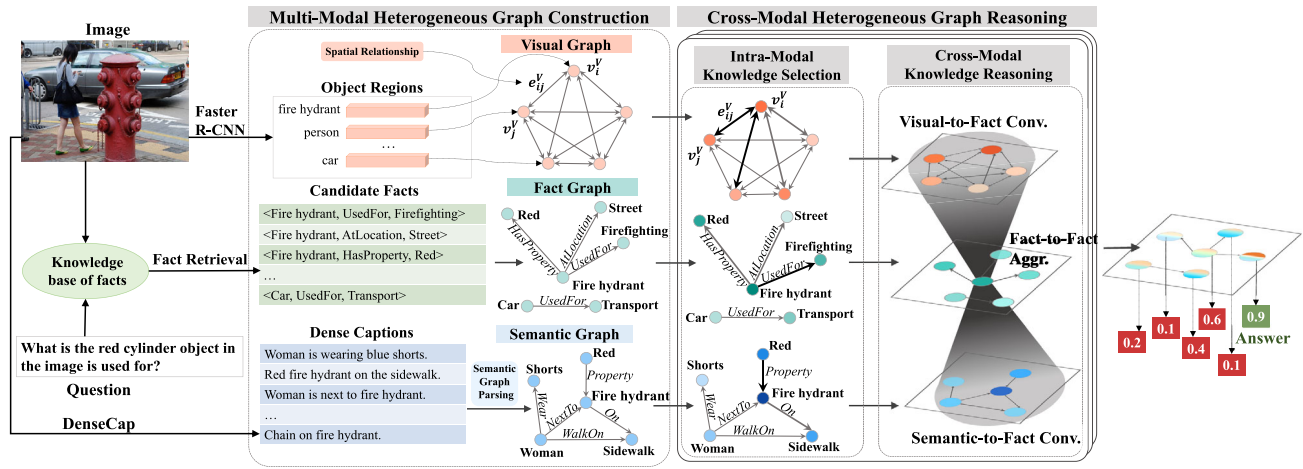

**Fig. 4** MUCKO architecture [124] (reused with permission). Best viewed in colour

porates knowledge from the visual graph. Another starts with the same knowledge graph concept but incorporates semantic knowledge. At the end of the recurrent operations, the outputs of the two pipelines are fused together with the question and original fact node. This fused feature is then used to predict the final answer. The change made to the cross-modal attention mechanism yields significant improvements in the FVQA benchmark when compared with MUCKO [124].

Liu et al. [126] also adopt a multi-modal approach, but use only the semantic and knowledge modalities. They propose a dual process system for FVQA that is based on the dual process theory from cognitive science [127]. Their approach utilises a BERT encoder to represent the input question and a Faster-RCNN [78]-based feature extractor to represent the image features. The first of the two systems, based on the Transformer architecture [40], joins these two representations into a single multi-modal representation. The second system then develops a semantic graph by turning dense region captions into textual scene graphs (using SPICE), as well as a knowledge graph generated using the question input. A message passing GNN is then used to identify the important nodes and aggregate information between them using an attention weighting. A joint representation for each knowledge graph node is then learned by combining the whole semantic graph with the node with relation to an attention weighting. This joint representation is then used to predict the final answer.

The GNN-VQA proposed in [128] makes use of a bidirectional GNN that fuses structured and unstructured multi-modal data through a 'supernode'. After extracting a semantic graph, they use a pretrained sentence BERT model to calculate the similarity between the potential answers and the Visual Genome region descriptions. The top-10 region descriptions in terms of similarity are averaged and used to define a concept node which is connected to each visual node of the semantic graph. Concepts are then extracted from the ConceptNet KG [67] using the labels of detected objects in the image and concepts extracted from potential answers. A GAT-based GNN is then used to construct better node representations which are then used to select the correct answer.

Moving away from the multi-modal approach, SGEITL [129] makes a semantic graph of the image and then follows Yang et al. [54] and introduces skip edges to the graph, essentially making it a complete graph. This graph then goes through a multi-hop graph Transformer, which masks the attention between nodes based on their distance, ensuring that only close by nodes are attended to. Through their work, they demonstrate that structural information is useful when approaching the complex VQA task.

With their TRiG model, Gao et al. [130] advocate taking an alternative approach to FVQA and rather than generat-ing the answer in some multi-modal space, they propose to use the textual space. They argue that this prevents further fusion with additional outside knowledge, and that as most of this data are in textual form, it makes sense to work in that domain. TRiG therefore has three components. It first converts the image into a caption using an off-the-shelf image captioning tool. The model then finds the top $K$ relevant facts from a knowledge base of Wikipedia articles before using a T5 backbone Transformer [131] to fuse and decode the <question, visual context, knowledge> triplet into an answer.

### 5.3 TextVQA

TextVQA is the sub-task of VQA where the answers require the model to be able to read text that appears in images. Typically this involves tasks like reading brand names from buildings or the title of book covers. This information can then be combined with an external knowledge base, enabling the models to answer questions such as 'Is the shop an American brand?' by reading the shop name and searching it in a knowledge base.

Gao et al. [132] focus on the in-image text and how it can be better leveraged to improve VQA. They use a novel multi-modal graph made up of fully connected visual, semantic, and numeric sub-graphs. Each sub-graph represents a unique modality that can be found in an image: visual entities (represented by image feature extractors), semantic meaning of discovered text (initially discovered by OCR), along with numeric values and their semantic meaning. The paper proposed a model that aggregates information across modalities together using a relevance score. Once the three modalities have been aggregated, an attention mechanism is deployed to help predict the final answer. The focus on different modalities proves a useful approach, with the model performing favourably in benchmarks (see Table 6).

Another work that makes use of multi-modal graphs is Liang et al. [133]. Their work uses both image features and scene text features (extracted by OCR) to generate a spatial relationship graph similar to that of [39]. The graph undergoes multi-head attention before being processed by a GNN that makes use of the attention weights. Multi-modal fusion is then used to join the node features with the question embedding and positional features. The output of this fusion operation is then used to predict a final answer.

## 6 Image retrieval

Image retrieval is the task of finding images from a database given some query. These queries can take many forms, including a similar image, a natural language query, or even a sketch. A common approach is to represent the database images as being in some space, where similar images are those with a minimal distance to the query. When this space is represented using graphs, GNNs become valuable for sharing features and acquiring more global context for the features.

Johnson et al. [38] show that a scene graph can be used as the input of the image retrieval system. By allowing end users to create a scene graph where nodes represent objects, attributes, and relationships, they are able to return appropriate images via a scene graph grounding process. This involves matching each scene graph object node with a bounding box predicted by an object detector, and is represented probabilistically using a conditional random field (CRF). The advantage of using scene graphs as search queries over natural language is that they scale well in terms of complexity. Once a basic scene graph has been constructed, it is straightforward for it to be extended and made more complex by adding additional nodes. Another advantage is that it reduces the operations required to map the search query to the image.

Following on from [38], Yoon et al. propose IRSGS (Image Retrieval with Scene Graph Similarity) [73], which makes use of a semantic graph. Given a query image, the model will generate a semantic graph and compare its similarity with graphs of images in the database. This graph comparison is achieved by taking the inner product of graph embeddings generated by a GNN (either GCN [134] or GIN[135]). One key contribution of the paper is the concept of Surrogate Relevance, which is the similarity between the captions of the images being compared. Surrogate Relevance is calculated using the inner product between Sentence-BERT embeddings of the captions. This measure is used as the training signal of the model to hone the feature embeddings generated by the GNN. The graph-to-graph comparison behind the model allows this work to better scale to large image databases when compared to [38]. The use of Surrogate Relevance allows the work to be potentially expanded to match against user queries if they are in the style of the captions used to power the relevance measure.

Using an image graph of $K$-nearest neighbour graph of images represented as feature embeddings, Liu et al. [90] propose using a GCN alongside a novel loss function based on image similarity. The feature embeddings are enhanced to account for a global context across the whole image database using a GCN. Similarity between images is calculated by taking the inner product of the feature embeddings. The higher the similarity, the better the retrieval candidate. The author's novel loss function is designed to move similar images closer together in the embedding space and dissimilar images further apart. Compared with [73], by using the inner product, the similarity measure is far more deterministic. However, unlike [73], it cannot be expanded to work alongside text-based image retrieval with a user query.

Zhang et al. [136] also use a $K$-nearest neighbour graph, but focus on improving the re-ranking process in content-

**Table 5** A table showing the model details and results of selected models trained and tested against the OK-VQA [51] and FVQA [50] datasets

| Model | Graphs Used | GNN Architecture | OK-VQA Top-1 Results | OK-VQA Top-3 Results | FVQA Top-1 Results | FVQA Top-3 Results |
|---|---|---|---|---|---|---|
| Out of the box [125] | Knowledge | GCN | – | – | 69.35 | 80.25 |
| Dual process [126] | Semantic, Knowledge | GCN | 29.43 | **32.83** | 63.57 | 76.47 |
| Mucko [124] | Visual, Semantic, Knowledge | GCN | – | – | 73.06 | 85.94 |
| GRUC [77] | Visual, Semantic, Knowledge | GCN | **29.87** | 32.65 | **79.63** | **91.20** |

Bold: Best score

**Table 6** A table showing the model details and TextVQA-Val Accuracy results of selected TextVQA models

| Model | Graphs Used | GNN Architecture | TextVQA-Val Accuracy (%) | TextVQA Test Accuracy (%) |
|---|---|---|---|---|
| MM-GNN [132] | Visual, Textual Semantic, Numeric | Multi-modal GNN | **31.44** | **31.10** |
| MCG [133] | Spatial | Multi-modal (contextual) GNN | 29.40 | 29.61 |

Bold: Best score

based image retrieval. A GNN is applied to aggregate features created from a modified adjacency matrix. Using a GNN allows the re-ranking process to de-emphasise nodes with a low confidence score.

Rather than use a pure $K$-nearest neighbour graph, the DGCQ model [137] is based on vector quantisation, a process from Information Theory for reducing the cardinality of a vector space. It can essentially be thought of as a many-to-one clustering technique where vectors in one space $\mathbf{x} \in \mathbb{R}^d$ are mapped to a single point in another space. A mapping function $q(\mathbf{x})$ maps the vector to a codeword $c_i$. These codewords make up a set of length $K$ known as a code book; thus, $q(x) \in \mathcal{C} = \{c_i; i \in \{0...(K-1)\}\}$. The learned code words are combined with image features to form landmark graph—based on the similarity graph except the graph also has nodes learned through the quantisation process. Once the landmark graph has been constructed, a GCN is use to propagate features with the objective of moving similar images closer together in the feature space. The use of vector quantisation allows for the landmark graph to exist in a lower-dimensional space, reducing computation when computing which images from the graph to return as candidates.

The authors of [74] move to adopt a multi-modal approach. They use GraphSAGE [99] to effectively learn multi-modal node embeddings containing visual and conceptual information from the connections in the graph. The distance between connected nodes are reduced, whilst the distance between disconnected nodes is increased. By using graph nodes that represent images as well as nodes representing meta-data tags, their model is able to provide content-based image retrieval as well as tag prediction. At inference time, images shown to the model can be attached to the graph through their $K$ nearest images, attached to relevant tags, or both. Unlike previous works [38, 73, 90], Misraa et al. [74] make use of multi-modal embeddings in the graph nodes.

Schuster et al. [82] continue the work of Johnson et al. [38], by creating a natural language parser that converts a query into a scene graph that can be processed by their work. This allows them to go beyond content-based image retrieval and move into text-based image retrieval. Their parser works by creating a dependency tree using the Stanford Dependency Parser [83] and then modifying the tree. They first execute a quantification modifier that ensures nouns are the head of the phrase. This is followed by pronoun resolution to make the relationship between two objects more explicit. Finally, plural nouns are processed. This involves copying noun instances when numeric modifiers are given. This textual scene graph is then mapped to images following [38].

Cui et al. [72] also tackle text-based image retrieval. They present work that makes use of a GCN to provide cross-modal reasoning on visual and textual information. Input features are split into channels which form a complete graph and undergo graph convolution. Once the textual and visual features are projected into a common space, they have their distances measured using the cosine similarity. These similarity scores are then stored in a matrix representing the similarities between visual and textual inputs.

Zhang et al. [138] tackle the challenging task of Composing Text and Image to Image Retrieval, where given a reference image and modification query the image retrieval system must find an image similar to the reference that contains the modifications outlined in the query. The principle challenge of this emerging task is its cross-modality nature. The authors tackle this challenge by first generating a spatial graph of the reference image and a textual feature of the modification query. These features are then concatenated before the graph is processed by a GAT whose attention mechanism has been altered to account for the directionality of the graph and the spatial data it encodes. A collection of GRUs that form a Global Semantic Reasoning GSR unit are then used to create the final embedding for the reference image. The same process is used on the target image but without the concatenation of the textual feature. A cross-modal loss function and adversarial loss function are combined to ensure that the features outputted by the Global Semantic Reasoning unit of the same category are moved closer together.

Chaudhuri et al. [93] adopt a Siamese-based network architecture where two similar inputs go into two separate networks that share weights. This network architecture typically uses contrastive loss or triplet loss to ensure the outputs of these networks are similar. The authors use a novel Siamese-GCN on a region adjacency graph that is formed by connecting adjacent segmented regions and weighting the edge accounting for the distance and angle between centroids of the regions. They apply their technique to high-resolution remote sensing images for content-based image retrieval. By using a Siamese-GCN with contrastive loss, the authors are able to learn an embedding that brings similar images together and forces dissimilar images apart. This work is then followed up by the authors in [139], where they add a range of attention mechanisms. They implement both node-level and edge-level attention mechanisms (in a similar style to GAT [102]). These attention mechanisms are then incorporated into the Siamese-GCN to yield improvements over their previous work.

Another work to incorporate a Siamese network design was Zhang et al. [140]. They use a three part network design to perform zero-shot sketch-based image retrieval with a Siamese-based encoding network which creates features of the image and associated sketch using ResNet50. These features are the concatenated together to create node features. The similarity between nodes is calculated using a metric function modelled by an MLP, and this operation is used to populate the adjacency matrix of a similarity graph. A GCN is then applied to the similarity graph to create fusion embeddings of sketch–image pairs. Rather than use an MLP to reconstruct the semantic information from the GCN embeddings, the authors chose to use a conditional variational autoencoder [141]. Doing so enables the model to generate semantic information for sketches of unseen classes, aiding the zero-shot component of the model.

## 7 Discussion and conclusion

In this section, we draw upon the views of Battaglia et al. [41], and discuss how the popular Transformer [40] can be viewed through the lens of GNNs. We then discuss how its dependence on consistent structure may pose challenges should image generation techniques be applied to create new training data for image captioning. The section concludes with a final summary of the paper and an overview of the challenges and future research directions that lie ahead for graph-based 2D image understanding.

### 7.1 Why GNNs when we have transformers?

Recent years have seen the rapid rise in popularity of the Transformer architecture [40]. Originally proposed in the Natural Language Processing domain, it was quickly applied as a generalised encoder in computer vision tasks [60]. Further work then expanded the architecture so that it can process images directly [142, 143], allowing it to operate as a backbone for common vision tasks. The wide range of applications the architecture can be applied to has led to it dominating much of deep learning in recent years.

There has been some effort by the community to unify the attention-based approach with GNNs. Battaglia et al. [41] proposes a more generic graph network which both Transformers and GNNs fall into. They present a viewpoint where Transformers can be viewed as a neural architecture operating on a complete graph.

Viewing GNNs and Transformers as graph networks shows that they share a number of similarities. Both architectures take a set of values and decide how much different values should be considered when transforming them to update the values, with GNNs ignoring nodes that are not connected and Transformers scaling the importance of an input. It is worth noting that if the graph being processed by a GNN is a complete graph, the graph network will allow all nodes to have their messages propagated to one being updated. Therefore, it is possible to view the Transformer as a special case GNN operating on a complete graph. Whilst GNNs use the read module to take advantage of an underlying structure, the Transformer learns one based on the task.

By applying a Transformer to a task, a graph structure is being learnt from scratch. Meanwhile, there are plenty of graph structures that appear naturally within Vision-Language tasks. This multitude of graph types allow for different structures to be taken for the image, from the semantic structure of an image to the hierarchical structure of the image with regards to the entire training set. Graphs appear naturally in the language component of the tasks as well, with sentence dependency trees being closely aligned to semantic graphs (when the semantic graph is made multi-partite as in the case of [80]). When clear graph representations of data exist, they should be utilised rather than ignored, instead of learning a graph structure using a more general purpose architecture. Utilising existing graph structures enables a graph network with the appropriate inductive biases to be deployed. It also results in fewer computations as messages are not being passed between all possible node connections. Looking at the results of various image captioning models (Table 3), it is clear that whilst the fully Transformer-based $\mathcal{M}2$ model performs impressively with a BLEU-4 of 39.1, models utilising GNN-based encoders outperform it. For example, the Dual-GCN [89] has a BLEU-4 of 39.7. Table 3 shows that the benefits of a Transformer lie in the language generation, rather than the image encoding. A GNN-based encoder seems to create a better representation of the image. This viewpoint is reinforced by the performance of [119]. Whilst their incorporation of geometric relationships into the Transformer attention model works well, it falls short of the performance of models that specifically use a spatial graph [87, 89, 115]. These results show that if a specific relationship, one that can be expressed explicitly as a graph, is being exploited by a model, then its architecture should make use of a GNN to take advantage of the graph. The spatial relationships between objects form a graph, and therefore, a GNN is well suited to make use of this information.

Although Transformers can be viewed as operating on a complete graph and pruning edges via the attention mechanism [41], we have demonstrated in this survey that this is not always the best approach. This raises the questions of 1) which graph should be used? and 2) How do we construct better graphs?

There are a wide number of aspects that researchers should consider when selecting a graph type for a GNN-based Vision-Language model. However, they all come down to experimentation and iteration during development. Whilst GCN-LSTM and VSUA demonstrate that a semantic graph is better suited to image captioning than a spatial graph, that is not to say spatial graphs no longer have a place in image captioning. It may be pertinent to incorporate a spatial graph if a model spatial reasoning is limited.

Whilst the challenge of producing better graphs is sometimes down to the optimisation of hyperparameters (in the case of the kNN-based image graph), better graphs are sometimes data or model dependent. Larger, more detailed knowledge graphs will yield improvements in fVQA and improvements in scene graph generation will produce richer semantic graphs. Scene graph generation is an incredibly vibrant field within Vision-Language and we direct readers to relevant surveys [144]. Better graph representations can be achieved through careful definition of the graph structure. Spatial relationships have been modelled in a number of different ways [39, 80, 119], leading to different model performance. Finally, developments in upstream tasks such as object detection can lead to better initial representation of visual components in the nodes of graphs. Advancements like VinVL [145] show that the use of more contemporary object detection techniques produces richer features that can be incorporated into the graphs used in Vision-Language tasks.

When it is possible to utilise multiple graphs, it is advantageous to do so when compared to using a single graph. As shown with image captioning (Table 3), architectures that only use a single graph type perform sub-optimally compared to their multi-graph counterparts. ARL [117], SUB-GC [79], and Topic [88] all use a single graph (spatial, semantic, and similarity, respectively) and all three suffer in benchmarks. Whilst Topic performs well in BLEU, METEOR, and ROGUE, when evaluated using metrics designed specifically for image captioning (SPICE and CIDEr) its performance falters against comparable models. This theme of multi-graph approaches performing more favourably is also found across the VQA, FVQA, and TextVQA tasks, with multi-graph approaches outperforming their single graph counterparts.

### 7.2 Latent diffusion and the future of image captioning

Currently, image captioning techniques are constrained by their training data. As popular as COCO is within the Computer Vision community for its wide ranging scenes and generalisability to the real world, it has its shortcomings. Captioning systems trained on it alone will never understand particular art styles, or objects outside of the 80 categories covered by the COCO dataset. The advent of image generation techniques such as DALLE·3 [146] present an opportunity for image captioning systems to go well beyond an 80 category limit and start understanding various stylistic elements of images. Work in this area is in its infancy [147, 148], but previous non-generative unsupervised approaches to image captioning are very promising [29].

We speculate that latent diffusion-based captioning may be a promising avenue of research. However, for this approach to work effectively, image generation techniques will need to develop further. Currently Stable Diffusion 2.1 [149] and similar systems do not understand structure as deeply as would be required for them to be able to replace the training data of a captioning system, although DALLE·3 [146] has shown improvements in this area. As impressive as they are, diffusion models can struggle to assemble images correctly when the prompt asks for something that is unlikely in real life. When asked to generate an image of 'a man giving a horse a piggyback ride', models such as DALLE·2 [150] and Stable Diffusion 2.1 [149] can sometimes struggle to *understand* the requested spatial relation between the two objects, resulting in the sample result shown in Fig. 5. Although DALLE·3 [146] produces an image that aligns to the basic prompt, when changing 'man' to 'woman', the produced image no longer matches the prompt as intended. It could be argued that so called 'prompt hacking' could coerce all the models to produce the desired relationship between the objects when said relationship is outside of common distributions. However, that argument fails to address the fact that these models fail to understand a relationship a toddler would understand.

Discovering examples of incorrect relationships in images is not just a case of dreaming up relationships between objects

**Fig. 5** A comparison of different text-to-image models when given a simplistic caption. Best viewed in colour

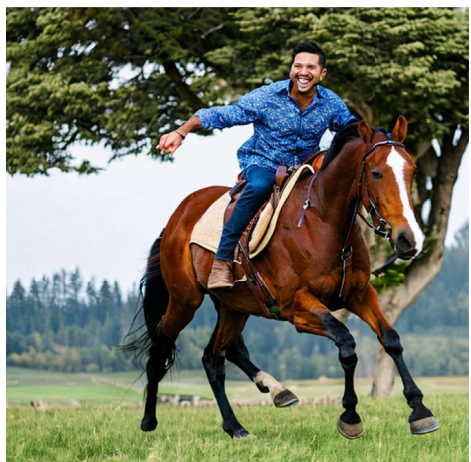

(a) Stable Diffusion [149] 2.1 with the prompt "a man giving a horse a piggyback ride"

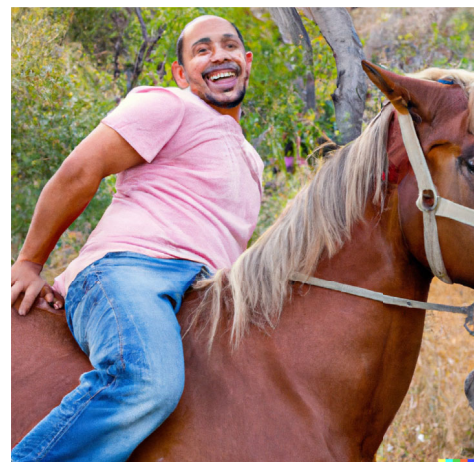

(b) DALLE·2 [150] with the prompt "a man giving a horse a piggyback ride"

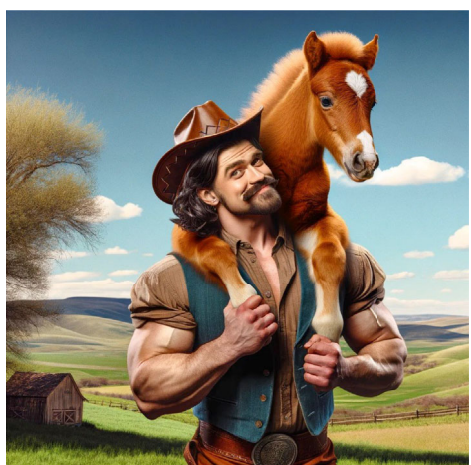

(c) DALLE·3 [146] with the prompt "a man giving a horse a piggyback ride"

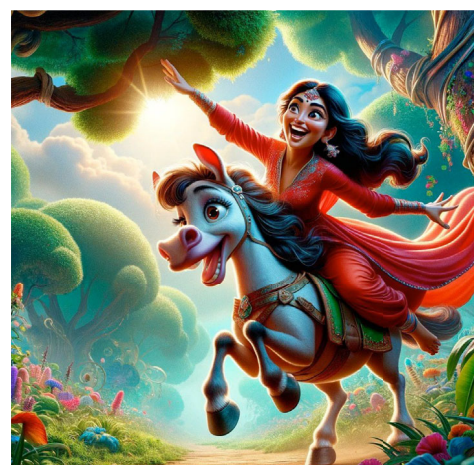

(d) DALLE·3 [146] with the prompt "a woman giving a horse a piggyback ride"

that are unlikely to exist in training data. Conwell and Ullman [151] conducted a participant study where they asked 169 people to select generated images that they felt well matched a given prompt. They found that across the generated images in their study, only 22% matched the original prompt. The authors conclude that *'current image generation models do not yet have a grasp of even basic relations involving simple objects and agents'* [151]. Whilst latent diffusion methods may play a role in the future of image captioning, they have a long way to go understanding structure before this is possible. In order for graph networks [41] to be applicable to diffusion generated training data, the structure within the image and the caption/prompt will need to be consistent. Supervised learning approaches require large amounts of very clean training data in order to work well, so graph networks [41] may struggle if the underlying structure in the image data is not as expected. However, we still expect that diffusion models will play a role within Vision-Language tasks going forward. The recent De-Diffusion Model [152] shows that the use of an image captioning model as the encoder portion of an autoencoder enables tasks such as VQA to be achieved by using a Large Language Model on the caption. This use of text as the latent representation of an image may provide a promising avenue of research.

### 7.3 Final notes

Vision-Language tasks such as image captioning and VQA pose significant opportunities for accessibility technology to be developed for those with sight impairment or severe sight impairment. Having widespread automatic alt-text generation on websites and applications enabling queries about images shared online, there is substantial impact that research in these fields can have. However, models trained on current datasets are prone to the biases of sighted humans. The questions asked in VQA datasets, and the captions given in image captioning datasets do not necessarily cater to the needs of possible end users of this technology. A lot is said in the field of the technology being applied to aid those with various levels of sight impairment, but little action is actually taken.

Whilst the release of trained models is promising, making these models available outside of the research community would be beneficial. Another direction the community could take is building on the work of VizWiz [153], a Vision-Language dataset curated for those with low vision. The dataset aims to highlight the accessibility requirements individuals with (severe) sight impairment require from models within this space. Whilst VizWiz [153] focuses on the VQA task, a similar route could be taken for image captioning, enabling researchers to tune models to ensure that the generated captions are useful to the people who need them the most.

The state of the art (SOTA) in Vision-Language tasks is currently dominated by large Transformer-based models developed by industrial labs [154–156]. This makes comparing these models to those discussed in this paper difficult given the model size and compute power used for training. However, there are a few take home points.

In the case of image captioning, the Transformer-based model $\mathcal{M}2$ is outperformed by GNN-based architectures, namely Dual-GCN [89]. This leads the authors to posit that there is a strong inductive bias in using imposed graph structures rather than allowing all relationships between detected objects to be processed using self-attention. The use of a global context graph (taking into account the whole dataset) alongside a local context graph (image-level relationships) by Dual-GCN [89] is shown to work extremely well and this dual graph approach could be the seed for future works.

It could be that given the scale of the models currently achieving SOTA that there are some emergent properties that develop in these models when they achieve such as scale. Future work should consider scaling graph-based architectures, such as those discussed in this survey, to the scale of the large models being produced by industry labs.

For FVQA and image retrieval, the graph-based approaches have stronger inductive biases for the reasoning stages of the tasks. Both tasks require the processing of graph data (in the case of a knowledge graph in FVQ or some graph representation of the search space in image retrieval). It is well documented that Transformers do not perform well on sparse graphs (such as knowledge graphs) or large graphs (such as those used in image retrieval).

The adoption of GNN-based image captioning techniques has proved promising. Given that this approach is relatively new, there is ample opportunity for further research to be carried out in this field. As shown in Sect. 4, the majority of image captioning techniques make use of either GCN or GGNN architectures. As GNNs develop and newer more expressive techniques are approached, the community should move to adopt these over traditional message passing style networks. Models such as GAT [102] may provide advantages over the techniques being used as they incorporate self-attention mechanisms into the architecture, a technique proved to yield impressive results given the popularity of the Transformer.

All the GNNs being used in the Vision-Language tasks discussed in the survey are built on the concept of homophily, i.e. similar nodes are connected by an edge. This is not always the case though given that a semantic graph connects dissimilar objects that are semantically related. Some of the graphs detailed are homophilic (e.g. image graph), but many others are not. This leads us to speculate that there are ample research opportunities for applying GNN architectures that respect the amount of homophily or heterophily of the graph being processed.

Another direction of research would be investigating combinations of different graph representations (at both the image level and dataset level) to identify combinations that work well together. Using different graph representations will allow for better utilisation of both local and global features.

The incorporation of outside knowledge into image captioning could provide an interesting research direction. It is often pointed out that image captioning is a useful accessibility technology for those with sight impairment. However, this assumes the user is an adult with a developed understanding of the world. Image captioning systems may struggle to be applied in a paediatric accessibility setting. Having the model explain the world in greater detail may be of use.

Another potential future research direction would be the unification of the three tasks discussed in this paper. Developing a single unified model that could perform competently in all three would hail an important breakthrough. In order to perform this, a model would have to have a common intermediary space for which it could map between the text and image spaces. We posit that this space would most likely be graph-based due to their expressive nature. However, a textual representation may also be performant as Gao et al. [130] showed reasoning in the text space improved performance over graph-based reasoning in VQA.

In summary, Vision-Language tasks such as those discussed in this paper are set to have a fruitful future, with many opportunities for various graph structures to be exploited.

## Declarations

**Conflict of interest** The authors declare no competing interests.

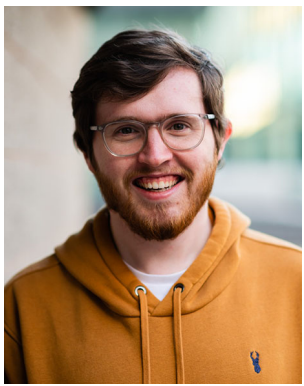

**Henry Senior** is a PhD student at Queen Mary, University of London in the Digital Environment Research Institute (DERI). His research focuses on the development and application of Graph Neural Networks to Vision-Language problems, such as image captioning. Before joining DERI, Henry completed a BSc(hons) in Computer Science with Professional Experience at the University of Salford and an MSc in Advanced Computer Science at the University of Liverpool. He also has three years of experience (full time and part time) as a Full Stack Software Engineer.

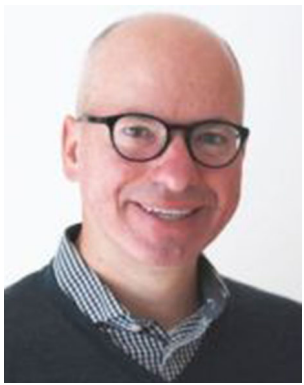

**Gregory Slabaugh** is Professor of Computer Vision and AI and Director of the Digital Environment Research Institute (DERI) at Queen Mary University of London. He is also Turing Liaison (Academic) for Queen Mary at The Alan Turing Institute. He earned a PhD in Electrical Engineering from Georgia Institute of Technology in Atlanta, USA. Previously, he was Chief Scientist in Computer Vision (EU) for Huawei Technologies R&D, and other prior appointments include City, University of London, Medicsight, and Siemens. He holds 38 granted patents and has approximately 200 per-reviewed publications. He regularly serves on the technical program committee for computer vision and machine learning conferences (CVPR, NeurIPS, AAAI) and related journals.

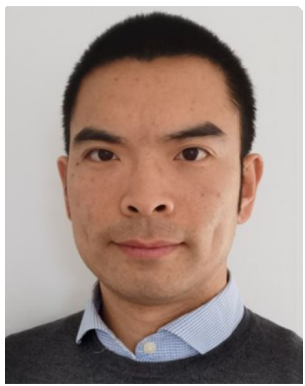

**Shanxin Yuan** is a Lecturer in Digital Environment at School of Electronic Engineering and Computer Science, Queen Mary University of London. Previously, he was a Senior Research Scientist in Computer Vision at Huawei Noah's Ark Lab, London Research Center, UK. The techniques he developed/involved have been shipped to several products. He received the PhD degree from Imperial College London, where he worked on hand pose estimation. His research interests are machine learning and computer vision, particularly 3D digital humans and computational photography. I regularly review for major computer vision conferences (CVPR, ICCV, ECCV, and NeurIPS) and related journals (TPAMI, IJCV and TIP).

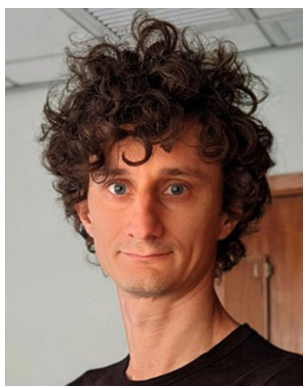

**Luca Rossi** received the PhD degree in Computer Science from Ca' Foscari University of Venice, Italy, in 2013. He is currently an Assistant Professor with The Hong Kong Polytechnic University, having held various positions with the University of Birmingham, Aston University, Southern University of Science and Technology, and Queen Mary University of London. He has published more than 60 papers in international journals and conferences. His research interests include the areas of Machine Learning, Data and Network Science. He is currently a member of the editorial board of the journal Pattern Recognition and vice-chair of the Technical Committee 2 of the International Association for Pattern Recognition.